\documentclass[10pt,twocolumn,letterpaper]{article}
\usepackage{cvpr}
\usepackage{times}
\usepackage{epsfig}
\usepackage{graphicx}
\usepackage{amsmath}
\usepackage{amssymb}

\usepackage{csquotes}
\usepackage{mathtools}
\usepackage[percent]{overpic}
\usepackage{siunitx}
\usepackage{subcaption}
\usepackage{verbatim}
\usepackage{xspace}
\newcommand{\refsec}[1]{Sect.~\ref{#1}}
\newcommand{\reffig}[1]{Fig.~\ref{#1}}
\newcommand{\reftab}[1]{Tab.~\ref{#1}}
\newcommand{\refequ}[1]{$\left(\ref{#1}\right)$}

\newcommand{\real}{\mathbb{R}}
\newcommand{\model}{\mathcal{M}}
\newcommand{\fmodel}{\mathcal{F}}
\newcommand{\object}{\model}
\newcommand{\scene}{\model'}
\newcommand{\fobject}{\fmodel}
\newcommand{\fscene}{\fmodel'}
\newcommand{\corrset}{\mathcal{C}}
\newcommand{\corrsetf}{\mathcal{C_\fmodel}}
\newcommand{\neighborset}{\mathcal{N}}
\newcommand{\voterset}{\corrset}
\newcommand{\voteset}{\Upsilon}
\newcommand{\globalsub}{G} % Subscript for global
\newcommand{\localsub}{L} % Subscript for local

\newcommand{\suchthat}{~:~}
\newcommand{\point}{p}
\newcommand{\feature}{f}
\newcommand{\pobj}{\point}
\newcommand{\pscn}{\point'}
\newcommand{\fobj}{\feature}
\newcommand{\fscn}{\feature'}
\newcommand{\T}{T}

\newcommand{\corr}{c}
\newcommand{\score}{s}
\newcommand{\threshold}{t}
\newcommand{\distance}{d}
\newcommand{\ethreshold}{\delta}
\newcommand{\compat}{\upsilon}
\newcommand{\prob}{\score}
\newcommand{\similarity}{\varsigma}
\newcommand{\kk}{\kappa}
\newcommand{\noise}{\sigma}

\newcommand{\precisionrecall}{recall \vs 1-precision\xspace}

\newcommand{\knen}{$\kk$-nearest Euclidean neighbors\xspace}
\newcommand{\approximately}{approx.\xspace}

\usepackage[pagebackref=true,breaklinks=true,letterpaper=true,colorlinks,bookmarks=false]{hyperref}

\cvprfinalcopy % *** Uncomment this line for the final submission

\def\cvprPaperID{1658} % *** Enter the CVPR Paper ID here

\ifcvprfinal\fi
\begin{document}

%%%%%%%%% TITLE
\title{In Search of Inliers: 3D Correspondence by Local and Global Voting}

\author{Anders Glent Buch$^1$ ~ Yang Yang$^2$ ~ Norbert Kr{\"u}ger$^1$ ~ Henrik Gordon Petersen$^1$\\
Maersk Mc-Kinney Moller Institute, University of Southern Denmark\\
{\tt\small $^1$\{anbu,norbert,hgp\}@mmmi.sdu.dk ~ $^2$yayan13@student.sdu.dk}}

\maketitle
%\thispagestyle{empty}

%%%%%%%%%%%%%%%%%%%% ABSTRACT %%%%%%%%%%%%%%%%%%%%
\begin{abstract}
We present a method for finding correspondence between 3D models. From an initial set of feature correspondences, our method uses a fast voting scheme to separate the inliers from the outliers. The novelty of our method lies in the use of a combination of local and global constraints to determine if a vote should be cast. On a local scale, we use simple, low-level geometric \emph{invariants}. On a global scale, we apply \emph{covariant} constraints for finding compatible correspondences. We guide the sampling for collecting voters by downward dependencies on previous voting stages. All of this together results in an accurate matching procedure. We evaluate our algorithm by controlled and comparative testing on different datasets, giving superior performance compared to state of the art methods. In a final experiment, we apply our method for 3D object detection, showing potential use of our method within higher-level vision.
\end{abstract}

%%%%%%%%%%%%%%%%%%%% INTRODUCTION %%%%%%%%%%%%%%%%%%%%
\section{Introduction}\label{introduction}
Consider the problem of matching two 3D point models $\object \subset \real^3$ and $\scene \subset \real^3$. For any point $\pobj \in \object$, the aim is to find the matching point $\pscn \in \scene$, if such a point exists. The result of this assignment is a \emph{correspondence}. When the full set of possible correspondences has been established, we say that $\object$ has been brought into correspondence with $\scene$. This represents a fundamental problem in computer vision and appears in \eg object detection.

\begin{figure}[t]
    \centering
    \begin{overpic}[width=0.475\linewidth]{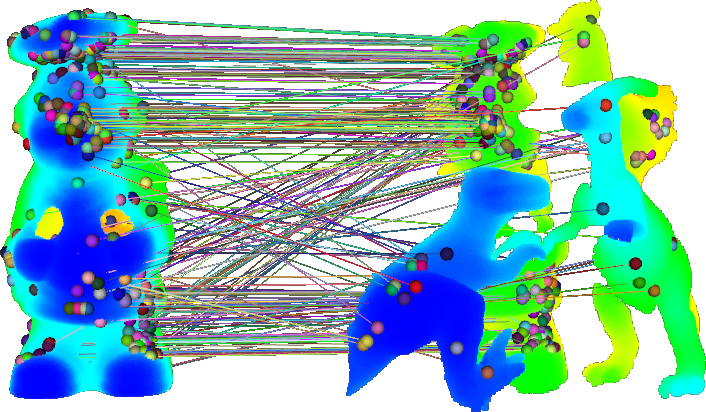}
        \put(7,-5){$\object$} % Below, object
        \put(70,-5){$\scene$} % Below, scene
    \end{overpic}
    \begin{overpic}[width=0.475\linewidth]{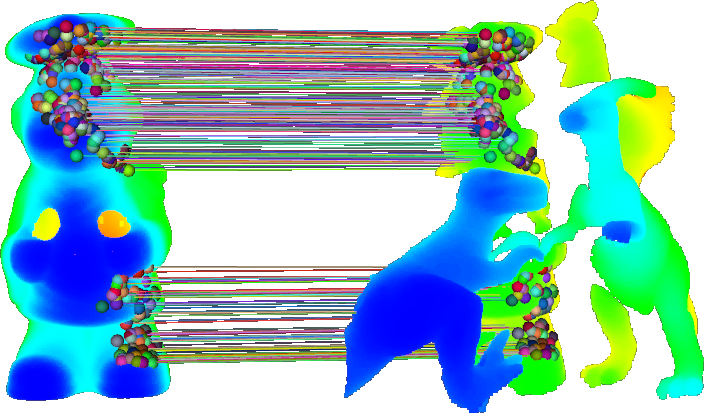}
        \put(7,-5){$\object$} % Below, object
        \put(70,-5){$\scene$} % Below, scene
    \end{overpic}
    \caption{Matching results between a complete 3D model $\object$ and a captured scene $\scene$ with cluttering objects and \SI{77}{\percent} occlusion of $\object$, taken from the dataset of \cite{mian2006}. Left: the \SI{1}{\percent} highest ranked correspondences obtained by Lowe's ratio criterion. Right: the \SI{1}{\percent} highest ranked correspondences after applying the proposed voting method.}
    \label{fig:corr_results_mian0}
\end{figure}

During local point matching or registration \cite{besl1992}, point assignments are made by spatial proximity of $\pobj$ and $\pscn$. Correspondences are progressively built by estimation of the relative transformation between $\object$ and $\scene$, followed by a reassignment. In this paper, the focus is on free-form matching problems, where no prior assumption can be made on the proximity of the models. To address this problem, local invariant features have been used extensively, both in images and in 3D \cite{bay2008,frome2004,johnson1999,lowe2004,mian2006,tombari2010}. In many practical scenarios, and especially in free-form matching problems, $\object$ and $\scene$ can be noisy and incomplete. In addition to this, either of the models can contain a significant amount of irrelevant data, or clutter. In \reffig{fig:corr_results_mian0}, we show an example of such a scenario, in which $\scene$ has been captured by a sensor. Although shape features can provide many good matches, one must expect a high amount of outliers due to repetitive structures, noise, clutter and occlusions.

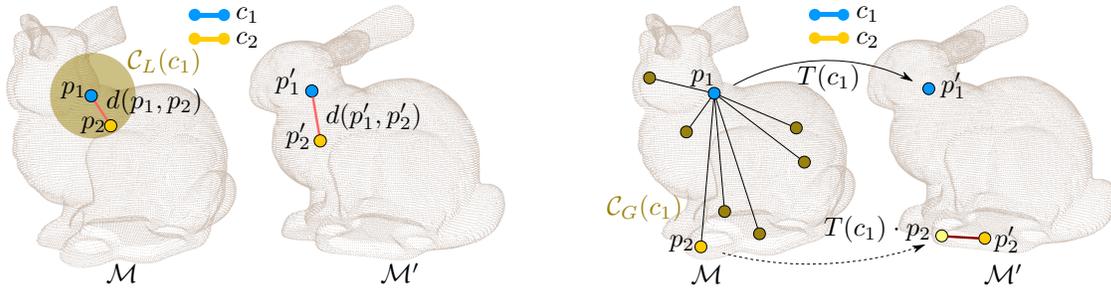
\begin{figure*}[t]
    \centering
    \begin{overpic}[height=0.2\linewidth]{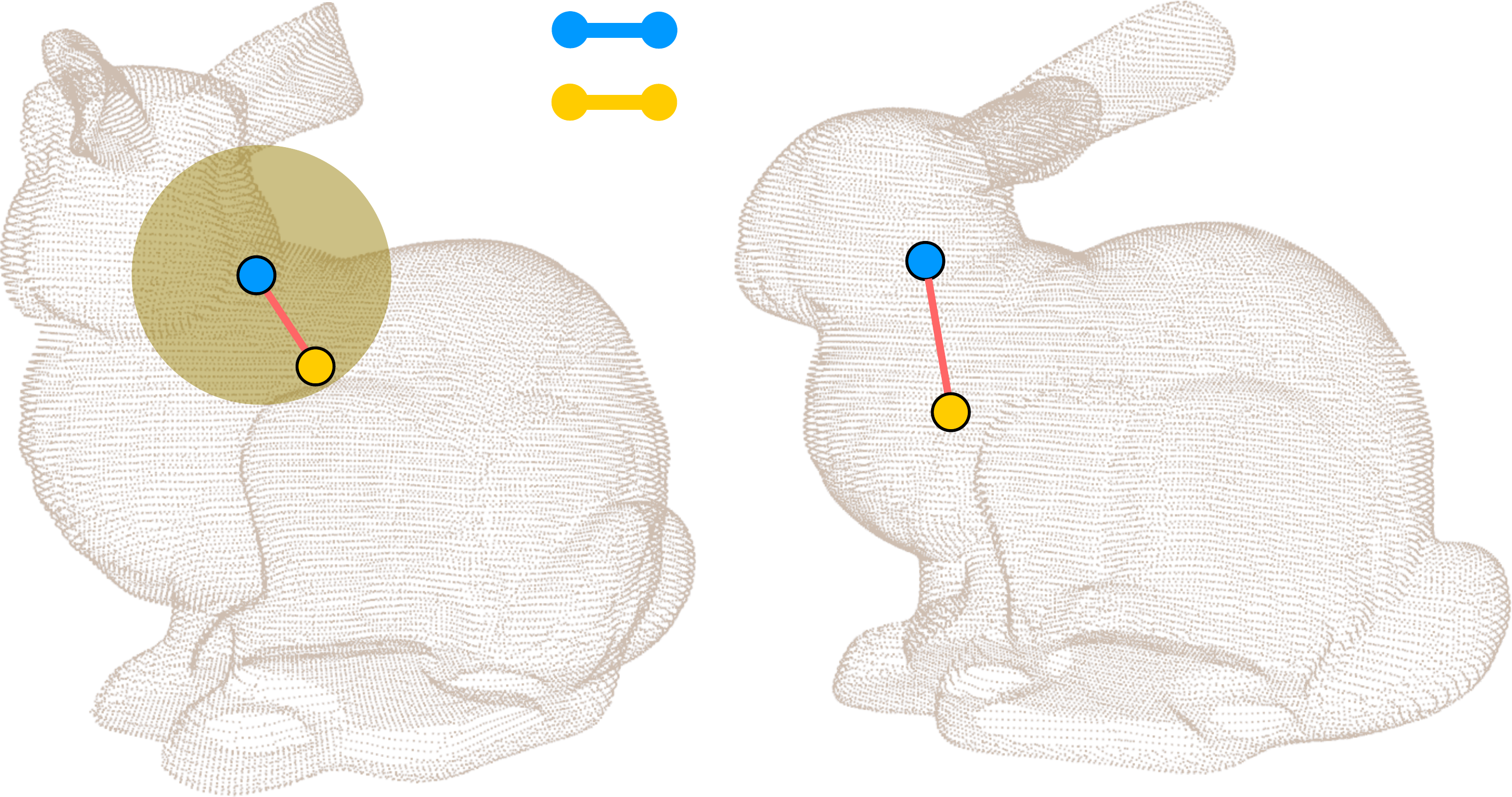}
        \put(20,-3){$\object$}
        \put(75,-3){$\scene$}
        \put(46,50){$\corr_1$}
        \put(46,45){$\corr_2$}
        % Left
        \put(23,40){ {\color[rgb]{0.61,0.51,0.05}$\corrset_\localsub(\corr_1)$} }
        \put(11,35){$\pobj_1$}
        \put(15,28){$\pobj_2$}
        \put(20,32){$\distance(\pobj_1,\pobj_2)$}
        % Right
        \put(54,36){$\pscn_1$}
        \put(56,25){$\pscn_2$}
        \put(64,29){$\distance(\pscn_1,\pscn_2)$}
    \end{overpic}
    \hspace{25pt}
    \begin{overpic}[height=0.2\linewidth]{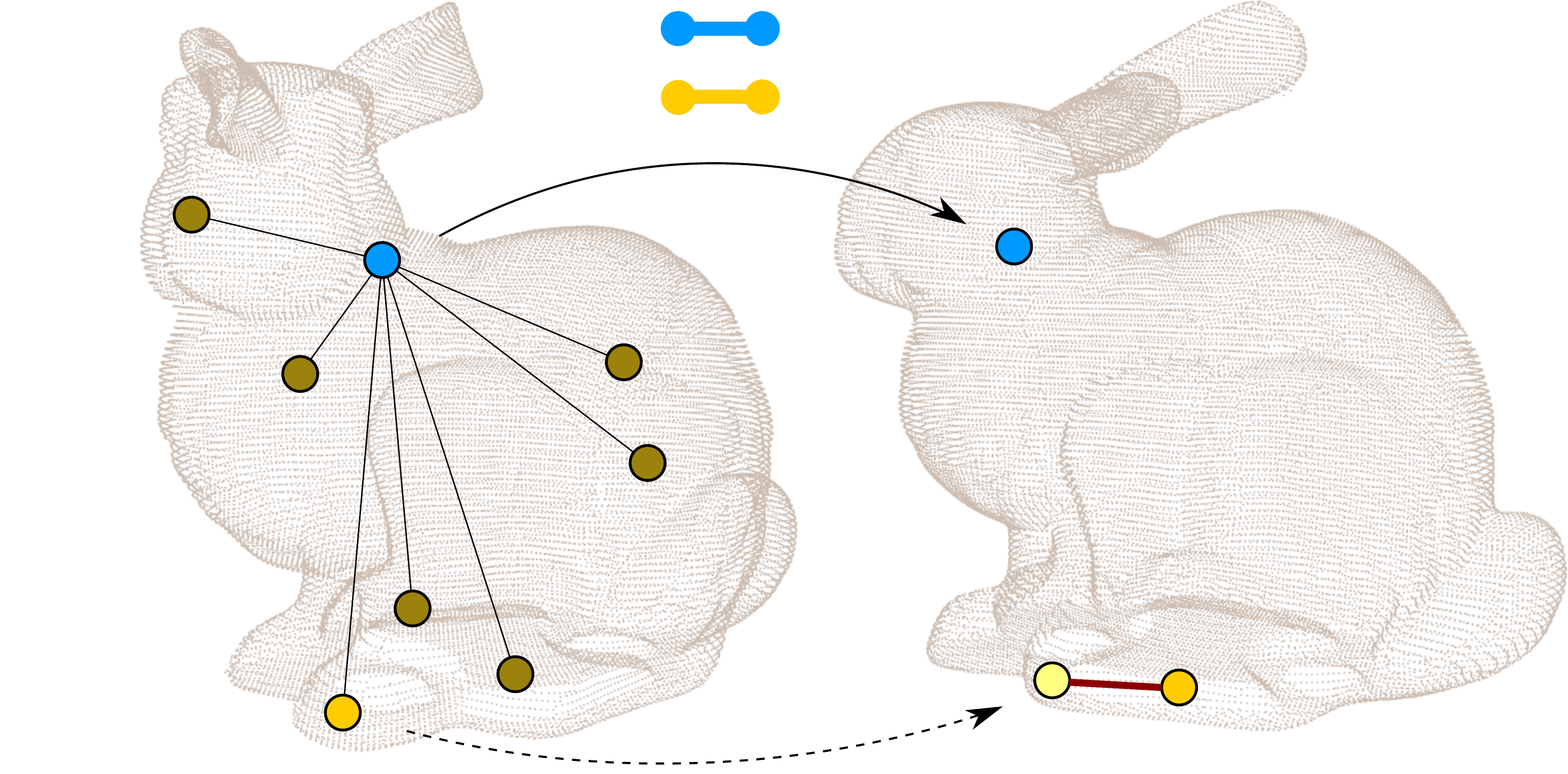}
        \put(20,-3){$\object$}
        \put(75,-3){$\scene$}
        \put(51,47){$\corr_1$}
        \put(51,42){$\corr_2$}
        % Left
        \put(3,10){ {\color[rgb]{0.61,0.51,0.05}$\corrset_\globalsub(\corr_1)$} } % Dark yellow
        \put(20,35){$\pobj_1$}
        \put(16,4){$\pobj_2$}
        % Right
        \put(40,34.5){$\T(\corr_1)$}
        \put(45,6){$\T(\corr_1)\cdot\pobj_2$}
        \put(67,33){$\pscn_1$}
        \put(77,4){$\pscn_2$}
    \end{overpic}
    \caption{Schematics of the entities involved in local (left) and global (right) voting. Left: local voters are collected in the spherical neighborhood of a correspondence (dark yellow circle). The invariant pairwise compatibility $\compat_\localsub$ is the minimum ratio of the light red distances. Right: global voters, from which $\corr_2$ is sampled, are located arbitrarily (dark yellow). The hypothesized transformation $\T$ (solid arrow) gives in this case an inaccurate alignment (dashed arrow) of $\pobj_2$ (light yellow). The covariant compatibility $\compat_\globalsub$ uses the dark red distance between the hypothesized point and the assigned point $\pscn_2$.}
    \label{fig:compat_schematic}
\end{figure*}

Our contribution is a method for finding correct correspondences within a set of initial or putative feature correspondences between two 3D models, corrupted by incorrect matches. Our method employs a two-stage voting procedure for estimating the likelihood of a correspondence being correct based on different pairwise constraints. At the first stage, we use a low-level invariant distance constraint imposed on the local neighborhood of each correspondence. At the second stage, we use the highest ranked correspondences of the first stage and enforce a covariant pairwise constraint. The first stage exploits the local dependency of correspondences. The second stage provides more independent observations and utilizes the fact that correct pose hypotheses are stable on a global scale. Our method efficiently finds correct correspondences, while rejecting outliers, giving an increase in matching precision. A visualization of the two constraint types is shown in \reffig{fig:compat_schematic} on the following page.

This paper is structured as follows. Related methods are outlined in \refsec{relatedwork}. Our method is presented in \refsec{proposedmethod}, and in \refsec{experiments} we provide experimental results. Finally, we draw conclusions and outline directions for future work in \refsec{conclusionsandfuturework}.

%%%%%%%%%%%%%%%%%%%% SCHEMATIC OF CONSTRAINTS

%%%%%%%%%%%%%%%%%%%% RELATED WORK %%%%%%%%%%%%%%%%%%%%
\section{Related work}\label{relatedwork}
Finding correspondences by voting processes has been subject to extensive investigation in the image domain. For completeness, we therefore start by outlining image-based correspondence finding methods, before describing the 3D methods used in our comparisons.

In \cite{leordeanu2005,torresani2008}, the correspondence problem between sparse sets of image features is cast to a graph matching problem. The former recovers the inliers through spectral analysis of the affinity matrix of the feature matches. The latter finds a solution by minimization of an objective function, taking into account both the appearance and spatial arrangement of feature points. Common to graph matching methods is a high degree of flexibility, allowing for non-rigid matching. However, the high computational complexity makes these methods infeasible for dense matching problems with a high cardinality (several thousand features). Additionally, the results reported in \cite{leordeanu2005,torresani2008} require very high inlier rates from the feature matches, which cannot be assumed for our application, as shown in \refsec{experiments}.

The \emph{pyramid match kernel} \cite{grauman2005} uses a fine to coarse matching strategy between sets of discriminative image features to achieve both robustness and discriminative power. Feature sets of unequal cardinality are allowed for, but accuracy is only justified for very sparse data (between 5 and 100 image features). In \emph{Hough pyramid matching} (HPM) \cite{tolias2011}, local covariant image feature correspondences are cast to a tessellated transformation space. Correspondences are afterwards rejected by pyramid matching. The \emph{Hough voting and Inverted Voting} (HVIV) \cite{chen2013} also uses covariant image features for casting votes, but preserves spatial locality of features before casting votes to arrive at accurate kernel densities. An additional inversion step propagates transformations on a local scale, by which an increased recall is achieved.

Lowe \cite{lowe2004} investigated the use of a simple, yet efficient method for selecting good SIFT feature correspondences. The method assigns a penalty equal to the \emph{ratio} of the closest to the second-closest feature distance. This measure is intrinsic to the feature space, and is immediately applicable to arbitrary feature types. On the other hand, only uniqueness is guaranteed, not necessarily discriminative power. Empirical data based on a large number of matched features suggest that this measure provides good separation of inliers and outliers. Even though the method has been used for image feature matches, it can be readily applied in 3D.

The \emph{geometric consistency} (GC) framework groups 3D correspondences into disjoint clusters, favoring correspondences likely to produce good transformation hypotheses. Initial work was done by Johnson and Hebert \cite{johnson1998,johnson1999}, where two oriented point pairs are grouped if they satisfy two geometric constraints. Firstly, the relative cylindrical coordinates of the point pair in first model must be compatible. Secondly, the method favors point pairs which have a large relative Euclidean distance, since this increases robustness of the subsequent transformation estimation step. Chen and Bhanu \cite{chen2007} relaxed the geometric constraints to only include a term which favors point pairs having similar distances. The omittance of orientation information in the grouping procedure increases robustness towards noise. Aldoma \etal \cite{aldoma2012} further robustified this method by applying a subsequent RANSAC \cite{fischler1981} step to remove spurious correspondences from each cluster. The GC methods apply constraints that are extrinsic to the feature space by using point entities pertaining to $\real^3$.

The method presented in this paper uses Lowe's feature distance ratio for initialization. At the first voting stage, we use a constraint based on Euclidean distance ratios, as opposed to the absolute distances used by the GC methods. In the second and final voting stage, we use covariant constraints similar to HPM and HVIV, but adapted to $SE(3)$. Unlike other methods, our method combines different constraints, both on a local and global scale, and uses a voting mechanism with downward dependencies.

%%%%%%%%%%%%%%%%%%%% METHOD %%%%%%%%%%%%%%%%%%%%
\section{Proposed method}\label{proposedmethod}
In this section, we describe our correspondence voting method. We start by introducing some terminology below, before describing our method in detail. To ease the readability of the following sections, we will refer to $\object$ as the \emph{object} and $\scene$ as the \emph{scene}.

\subsection{Terminology}
In the following, a correspondence $\corr$ between the object and scene models $\object$ and $\scene$ is parameterized by two matched points $\pobj\in\object$, $\pscn\in\scene$ and a real-valued matching score $\score$:
\begin{equation}
    \corr=\left(\pobj,\pscn,\score\right)
\end{equation}
Denote the feature space associated with $\object$ and $\scene$ as $\fobject$ and $\fscene$, respectively. A feature is computed for each point, \ie the feature sets are equivalent to the model point sets and have the same order. A feature-based correspondence is obtained by matching an invariant feature vector $\fobj \in \fobject$ with the nearest matching feature $\fscn \in \fscene$. Denote the general $n$-dimensional Euclidean $L_2$ distance metric as $\distance$:
\begin{equation}
    \distance(a,b) \coloneqq \|a-b\|_{L_2} \qquad a,b\in\real^n
\end{equation}
Associating a score to the feature match can now be done by the negative of the matching distance:
\begin{equation}
    \score_\fmodel(\corr) \coloneqq -\distance(\fobj,\fscn)
    \label{eq:scoredistance}
\end{equation}
The set of all correspondences is denoted $\corrset = \object \times \scene \times \real$. For free-form correspondence matching problems, there exists a unique subset of \emph{correct} correspondences that brings $\object$ into correspondence with $\scene$:
\begin{equation}
    \corrset_{Correct} \subset \corrset
\end{equation}
which represents the objective of the matching process.

In the case of occlusions (as in \reffig{fig:corr_results_mian0}), some points do not have a correspondence, and the scene $\scene$ contains an incomplete instance of $\object$. However, when performing dense feature matching---as we do in this work---all features in $\object$ will be assigned a \emph{putative} feature match in $\scene$ by the feature matching process. We denote this initial set $\corrsetf \subset \corrset$, and this serves as the input to our method. The problem is now to find the correspondences in $\corrsetf$ that are also part of $\corrset_{Correct}$ (the inliers), while rejecting those that are not (the outliers). This makes the problem of finding correspondence a binary classification problem.

The \emph{recall} of a matching method is defined as the ratio of correctly accepted correspondences to the number of inliers in $\corrsetf$. The \emph{precision} of a method is the ratio of correctly accepted correspondences to the total number of accepted correspondences. The initial \emph{inlier fraction} in $\corrsetf$ is thus the precision of the feature matching. An accurate matching method should therefore accept as many of the inliers as possible, while rejecting as many outliers as possible, giving an increase in precision.

\subsection{Overview}
The basic assumption behind our approach is that within the complete set of input feature correspondences $\corrsetf$, the inliers should systematically satisfy certain geometric constraints, while the outliers should only do so randomly. We enforce these constraints in a voting framework where each correspondence is paired with a number of voter correspondences. Each positive vote increases the likelihood or ranking score of a correspondence. We bootstrap the process by an initialization step based on the feature distance ratio, and then perform two voting stages. At the first stage, we use invariant distance constraints on a local scale by collecting voters in the immediate neighborhood of each correspondence. The fraction of positive votes gives a crude ranking of each correspondence. In the second voting stage, we find voters on a global scale, based on the first stage ranks, and enforce a covariant constraint. We introduce an additional dependency between the stages by accumulating all votes.

\subsection{Initialization}
Our method requires an initial ranking of the input correspondences. We start by ranking all input correspondences by the feature distance ratio, which has proven more discriminative than the closest feature distance. Lowe's ratio penalizes correspondences by the ratio of the closest to the second-closest matching feature distance. Since $\distance$ is a metric, this ratio will always lie in the interval $\left[0,1\right]$, and we can define the ranking score of a feature correspondence as:
\begin{equation}
    \score_{Ratio}(\corr) \coloneqq 1 - \frac{\distance(\fobj,\fscn_1)}{\distance(\fobj,\fscn_2)}
    \label{eq:scoreratio}
\end{equation}
where $\fscn_1$ and $\fscn_2$ are used for denoting the closest and second-closest feature match of $\fobj$, respectively. The ratio method then performs hard thresholding as follows:
\begin{equation}
    \corrset_{Ratio} = \lbrace \corr \in \corrsetf \suchthat s_{Ratio}(\corr) \geq \threshold_{Ratio} \rbrace
    \label{eq:corrsetratio}
\end{equation}
In the original work, an upper threshold for the penalty of 0.8 was determined using empirical data, giving a lower threshold on the score of $\threshold_{Ratio} = 0.2$.

\subsection{First voting stage: local invariants}
At the first voting stage, we locate the \knen $\neighborset$ of each correspondence on the object. For each correspondence $\corr$, we thus get a subset $\neighborset(\corr) \subset \corrsetf$ with $\left|\neighborset(\corr)\right| = \kk$. The number of neighbors $\kk$ is a free parameter of our method, and specifies the \emph{sample size}. We can now collect local \emph{voters} $\voterset_{\localsub}$ as the subset of neighbors that satisfy the ratio threshold \refequ{eq:corrsetratio}:
\begin{equation}
    \voterset_{\localsub}(\corr) = \lbrace \neighborset(\corr) \cap \corrset_{Ratio} \rbrace
    \label{eq:voterlocal}
\end{equation}
We pair the correspondence with each voter neighbor and measure their compatibility $\compat_{\localsub}$ by the minimum ratio of Euclidean distances between the object points and the corresponding scene points (see \reffig{fig:compat_schematic}, left):
\begin{equation}
    \compat_{\localsub}(\corr_1,\corr_2) \coloneqq \min \left( \frac{\distance(\pobj_1,\pobj_2)}{\distance(\pscn_1,\pscn_2)}, \frac{\distance(\pscn_1,\pscn_2)}{\distance(\pobj_1,\pobj_2)} \right)
    \label{eq:complocal}
\end{equation}
where the minimum of the two possible ratios is taken to get a result in $\left[0,1\right]$. By using a relative distance ratio in $\compat_{\localsub}$, the compatibility function becomes invariant to the absolute sizes of the involved distance pairs.% TODO: Is this even desirable?

The set of positive local \emph{votes} $\voteset_\localsub$ is the subset of local voters with a high compatibility:
\begin{equation}
    \voteset_{\localsub}(\corr) = \lbrace \corr_L \in \voterset_{\localsub}(\corr) \suchthat \compat_{\localsub}(\corr,\corr_L) > \similarity \rbrace
\end{equation}
where $\similarity \in \left[0,1\right[$ is the lower \emph{similarity} and is the second free parameter of our method. Larger values make the method more restrictive (giving fewer votes), but this also requires a more accurate representation of the surfaces to be reliable.

Finally, the likelihood, or estimated local score, $\prob_{\localsub}$, giving evidence of a correspondence under the local constraint, is calculated as the ratio of votes to the number of voters:
\begin{equation}
    \prob_{\localsub}(\corr) = \frac{\left| \voteset_{\localsub}(\corr) \right|}{\left| \voterset_{\localsub}(\corr) \right|}
    \label{eq:problocal}
\end{equation}

It is worth noting that the local voter collection process \refequ{eq:voterlocal} in some cases returns very small ($\ll\kk$) or even empty sets, depending on how many neighbor features pass the ratio test. Although this rarely happens in our experience, we must handle this by setting $\prob_{\localsub} = 0$ in the case of an empty set. The second voting stage explicitly handles the case of small voter sets by accumulating votes, as described in the following.

\subsection{Second voting stage: covariant surface points}
As previously mentioned, the first voting stage produces a crude estimate of each correspondence being correct using local neighbors in $\object$. This is justified by other studies, which have verified that correspondences exhibit local dependencies \cite{chen2013,yarlagadda2010}, meaning that correct correspondences often occur together. However, this also implies that inliers occurring near outliers passing the ratio test will get a low local score. We address this issue by introducing a second voting stage where the $\kk$ globally highest ranked correspondences of the first stage are used.

We start by reordering $\corrsetf$ according to \refequ{eq:problocal} to get a monotonically decreasing sequence in $\prob_{\localsub}$, denoted $\corrset_{\score_{\localsub}}$. We take out the $\kk$ top ranked correspondences, and arrive at a set of feasible voter correspondences for use in the global stage:
\begin{equation}
    \voterset_{\globalsub} = \lbrace \corr_i \in \corrset_{\score_{\localsub}} \rbrace_{i=1}^\kk %\qquad \left|\voterset_{\globalsub}\right| = \kk
    \label{eq:voterglobal}
\end{equation}
Since sampling is now based on $\score_\localsub$, the voters are collected globally on $\object$. We have also tested using a different number of global voters than $\kk$ at this stage, but found that best performance was achieved by reusing $\kk$. Unlike the local stage, all correspondences share the same voters, and we always have $\left|\voterset_{\globalsub}\right| = \kk$. We now compute a hypothesis transformation $\T \in SE(3)$ for each input correspondence in $\corrsetf$ using the reference frame (RF) associated to each feature point:
\begin{equation}
    \T(\corr) = \T(\pscn)^{-1} \cdot \T(\pobj)
\end{equation}
The use of RFs is common in images \cite{bay2008,leutenegger2011,lowe2004,rublee2011}, where the local RF consists of a pixel position, an orientation angle and a scale. Recently, methods for finding repeatable RFs for 3D shape features have emerged \cite{mian2006,tombari2010}, and we require this information to be available.% We also tested using oriented surface point pairs \cite{drost2010} for estimating $\T$, but found this to be both slower (since RFs now had to be computed using all possible pairs) and more inaccurate due to the magnified noise of the relative surface normal angles.

The transformation $\T$ gives a hypothesis pose for bringing $\object$ into correspondence with $\scene$. Two correspondences $\corr_1$ and $\corr_2$ are compatible if $\corr_2$ \emph{covaries} with the transformation hypothesized by $\corr_1$. We thus arrive at the following global compatibility function $\compat_{\globalsub}$ (see \reffig{fig:compat_schematic}, right):
\begin{equation}
    \compat_{\globalsub}(\corr_1,\corr_2) \coloneqq \distance \left( \T(\corr_1)\cdot\pobj_2 , \pscn_2 \right)
\end{equation}
We now find global votes $\voteset_{\globalsub}$ by applying both the local and the global constraint to the global voters $\voterset_\globalsub$:
\begin{equation}
    \voteset_{\globalsub}(\corr) = \lbrace \corr_G \in \voterset_{\globalsub} \suchthat \compat_{\localsub}(\corr,\corr_G) > \similarity \wedge \compat_{\globalsub}(\corr,\corr_G) < \ethreshold \rbrace
\end{equation}
where $\ethreshold$ is a Euclidean distance tolerance. To compensate for noise and inaccuracies in the RF rotation estimation, we set this tolerance to five times the point cloud resolution. If the resolution is not known a priori, it is estimated as the median distance between any model point and its nearest Euclidean neighbor. The local constraint $\compat_\localsub$ is enforced on the global voters for two reasons. Firstly, the distance ratio constraint should be satisfied for rigid objects, no matter if correspondences are paired locally or globally. Secondly, $\compat_{\localsub}$ is computationally cheap, and thus serves as a prerejection step to the more expensive $\compat_{\globalsub}$.

We integrate all votes and arrive at the final score function $\score$ as the likelihood computed by accumulating both local and global votes:
\begin{equation}
    \prob(\corr) = \frac{ \left|\voteset_{\localsub}(\corr)\right| + \left|\voteset_{\globalsub}(\corr)\right| }{ \left|\voterset_{\localsub}(\corr)\right| + \left|\voterset_{\globalsub}(\corr)\right| }
\end{equation}
which also makes it clear how small local voter sets is handled: smaller number of local voters gives higher relative importance to the global voters, which is a desirable effect as it reduces the bias from the small local sample size. We stress that in both stages the computed likelihoods have a downward dependency on the previous stage, introduced by the voter selection processes \refequ{eq:voterlocal} and \refequ{eq:voterglobal}. This guided sampling increases precision, as we will demonstrate in \refsec{experiments}.

\subsection{Thresholding}
Here we shortly describe how we perform the final thresholding to separate the inliers from the outliers based on the computed scores. The function $\score(\corr)$ is real-valued, so the problem is now to calculate a decision threshold $\threshold \in \left[0,1\right]$, based on some optimality criterion.

To address this, we apply a well-known method from the image processing domain, namely Otsu's adaptive thresholding method \cite{otsu1979}, which is non-parametric and finds the optimal decision threshold in a sampled univariate distribution under the assumption of bimodality. The method estimates the probability density function of the data by a histogram, and then exhaustively searches for the threshold which maximizes the between-class variance. In \refsec{experiments} we give experimental justification for the use of this method.

\subsection{Computational considerations and complexity}
We end the description of our approach by considering its time complexity. The feature estimation process requires local neighbors, and it is often possible to reuse these point neighbors at the local stage of our method. If not, neighbors can be found in logarithmic time by spatial indexing, such as $k$-d trees. In all experiments presented below, we have reused the feature neighbors for collecting local voters.

Since we use a fixed sample size $\kk$, and these are collected on the object, our algorithm is linear in the number of object points. The three components of our algorithm (initialization, local and global voting) each require a loop over all input correspondences, where the local and global stages each have an upper operation count of $\kk$ per correspondence. We thus get a final time complexity of $O(\left|\object\right| + \kk\cdot\left|\object\right| + \kk\cdot\left|\object\right|) = O(\left|\object\right|)$.

%%%%%%%%%%%%%%%%%%%% EXPERIMENTS %%%%%%%%%%%%%%%%%%%%

%%%%%%%%%%%%%%%%%%%% NOISY BUNNYS
\begin{figure}[t]
    \centering
    \begin{subfigure}[c]{0.2375\linewidth}
        \includegraphics[width=\linewidth]{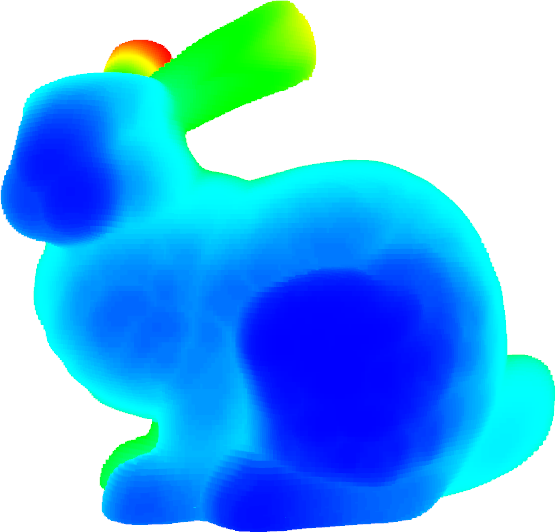}
    \end{subfigure}
    \begin{subfigure}[c]{0.2375\linewidth}
        \begin{overpic}[width=\linewidth]{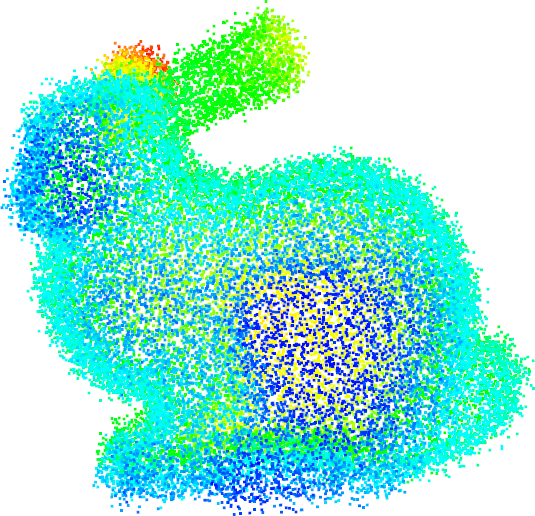}
            \put(10,-15){\footnotesize{$\noise = \SI{2.5}{\milli\meter}$}}
        \end{overpic}
    \end{subfigure}
    \begin{subfigure}[c]{0.2375\linewidth}
        \begin{overpic}[width=\linewidth]{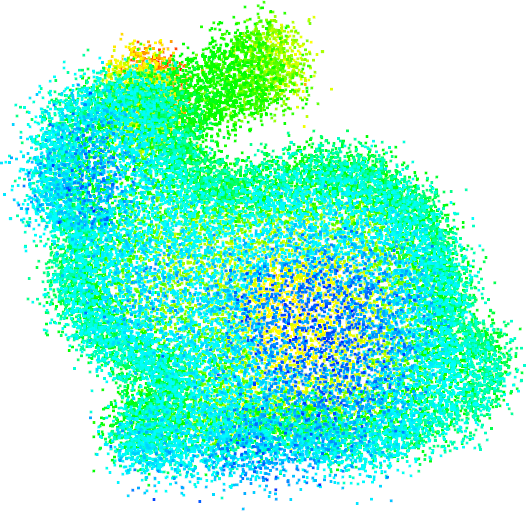}
            \put(10,-15){\footnotesize{$\noise = \SI{5.0}{\milli\meter}$}}
        \end{overpic}
    \end{subfigure}
    \begin{subfigure}[c]{0.2375\linewidth}
        \begin{overpic}[width=\linewidth]{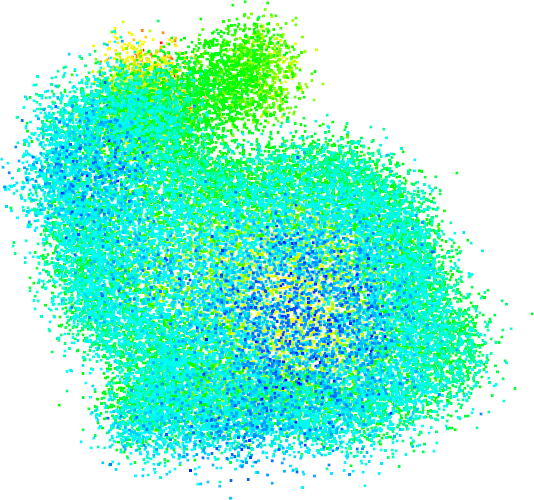}
            \put(10,-15){\footnotesize{$\noise = \SI{7.5}{\milli\meter}$}}
        \end{overpic}
    \end{subfigure}
    \caption{Original version of the Stanford \emph{Bunny} model (leftmost) and three of the test models used in the controlled experiment, color rendered by depth value.}
    \label{fig:bunny}
\end{figure}

\section{Experiments}\label{experiments}
We have performed both controlled and comparative experiments to evaluate our method. The standard evaluation procedure for feature matching is \precisionrecall \cite{mikolajczyk2005}, and we adopt these measures here. All methods are evaluated by varying the threshold $\threshold$ on the score associated with the method. In addition, we compute maximum $F_1$ scores, giving a conservative estimate of the overall accuracy.

With regard to the feature estimation, there are many variabilities, \eg support radius, matching distance metric \etc. We found that changing the feature or the radius has minor impact on relative performances. Like similar studies in the image domain \cite{grauman2005,tolias2011}, we use the same feature in all tests, the SHOT feature \cite{tombari2010}, which can be regarded as a SIFT-like shape feature. The SHOT features provide RFs for use in our global voting stage. The radius is set to \SI{0.015}{\meter} for a good trade-off between robustness and discrimination. For neighbor search, we use $k$-d trees \cite{muja2009} to locate both point and feature neighbors by the $L_2$ metric.

All experiments have been performed in a single-threaded C++ application using a laptop computer equipped with a \SI{2.2}{\giga\hertz} processor and \SI{8}{GiB} memory.

\subsection{Controlled experiment}
We start with a controlled experiment in order to test the effect of both noise and parameter changes on our method. All experiments described here are performed using the full Stanford \emph{Bunny} model,\footnote{\url{http://graphics.stanford.edu/data/3Dscanrep}} which contains 35947 vertices. We use the vertices and normals of the original mesh and add isotropic Gaussian point noise of increasing standard deviation $\noise = \lbrace 0.5, 1.0, \ldots, 7.5\rbrace~\si{\milli\meter}$, before computing features. For the high noise values, all local structures are severely distorted, rendering feature matching very challenging. See \reffig{fig:bunny} for an illustration.

\begin{figure}[t]
    \centering
    \begin{subfigure}[c]{0.475\linewidth}
        \includegraphics[width=\linewidth]{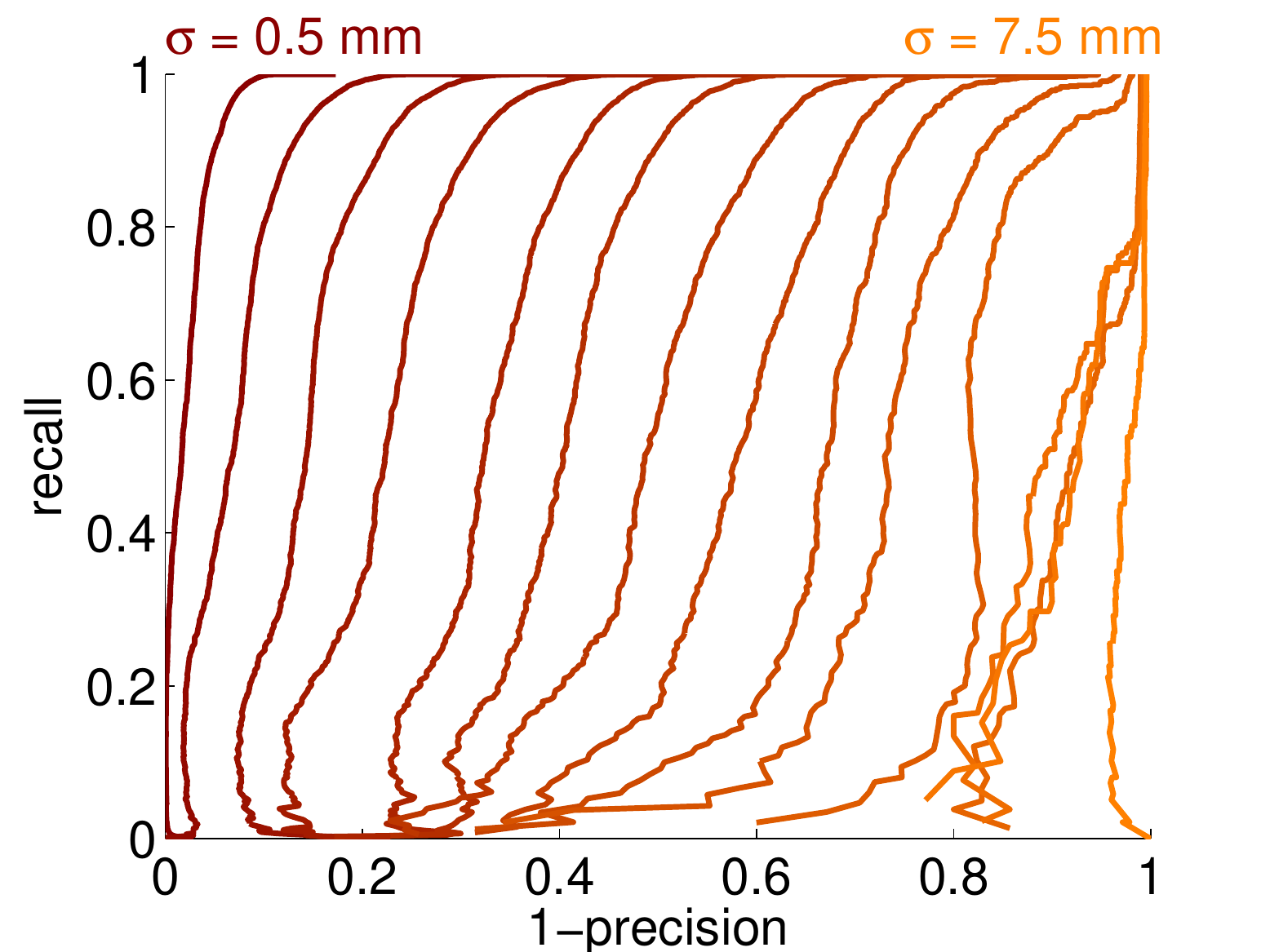}
    \end{subfigure}
    \begin{subfigure}[c]{0.475\linewidth}
        \includegraphics[width=\linewidth]{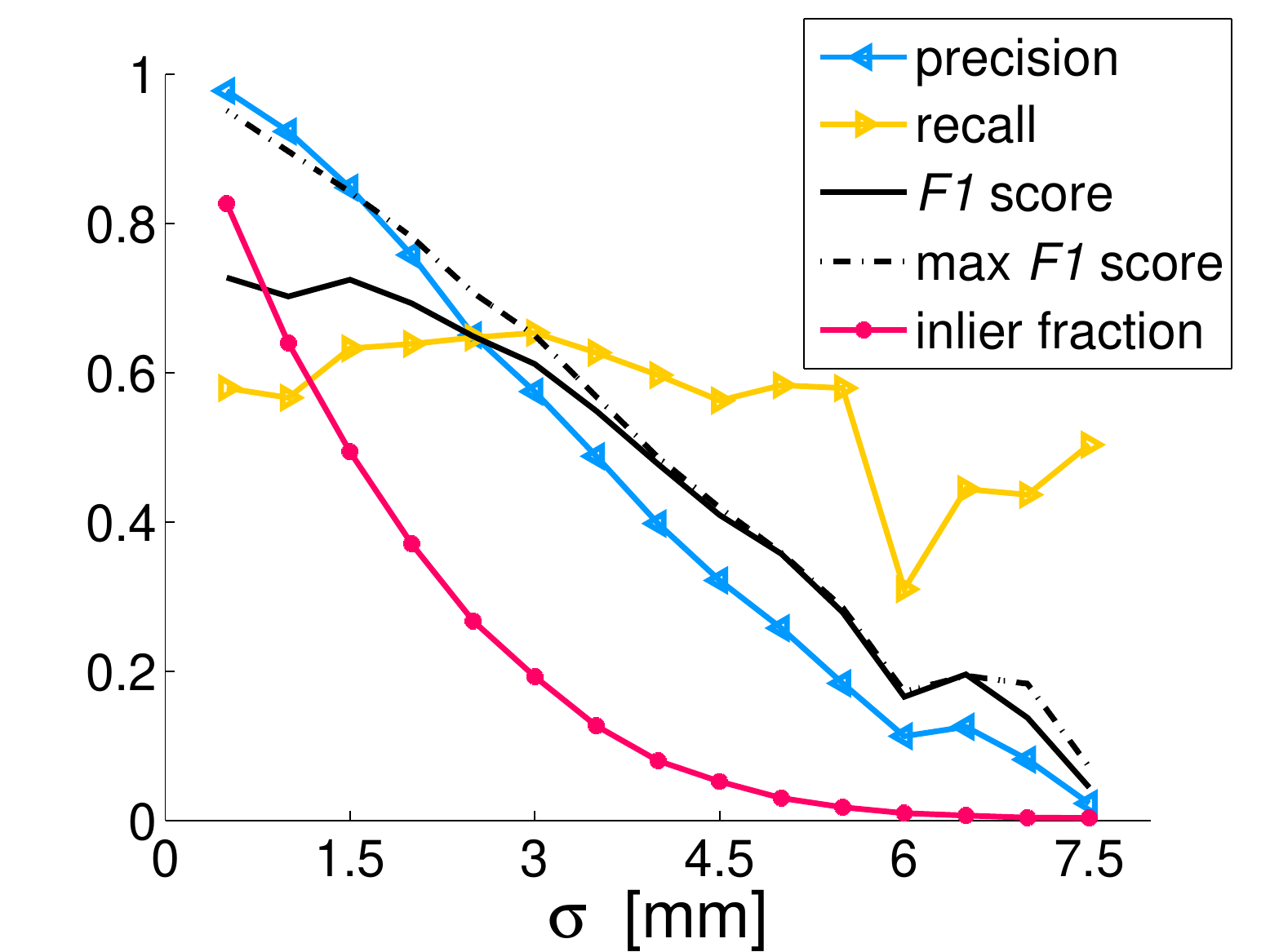}
    \end{subfigure}
    \caption{Performance measures for the \emph{Bunny} experiment for increasing noise. Left: \precisionrecall curves for all 15 noise levels (dark red to orange, left to right). Right: Precision, recall and $F_1$ score at decision threshold, maximum possible $F_1$ score and inlier fraction.}
    \label{fig:output_bunny_noise}
\end{figure}

The original noise-free model is paired with each of the 15 noisy versions, and we evaluate our method on all 15 shape pairs in order to measure the degradation in performance as a result of noise. The results of this experiment can be seen in \reffig{fig:output_bunny_noise}, showing both \precisionrecall curves for each noise level and performance measures at the decision threshold determined by our method.

%%%%%%%%%%%%%%%%%%%% BUNNY RESULTS, NOISE LEVELS

The results in \reffig{fig:output_bunny_noise} show a close to linear drop in precision with increasing noise, while achieving an almost constant recall, even though there is a rapid, sublinear drop in the inlier fraction. The $F_1$ score at the decision threshold $\threshold$ is close to the optimal value, which indicates good classification accuracy of the thresholding method. The matching destabilizes for $\noise \geq \SI{6.0}{\milli\meter}$, where the inlier fraction is close to zero.

\begin{figure}[t]
    \centering
    \begin{overpic}[width=0.475\linewidth]{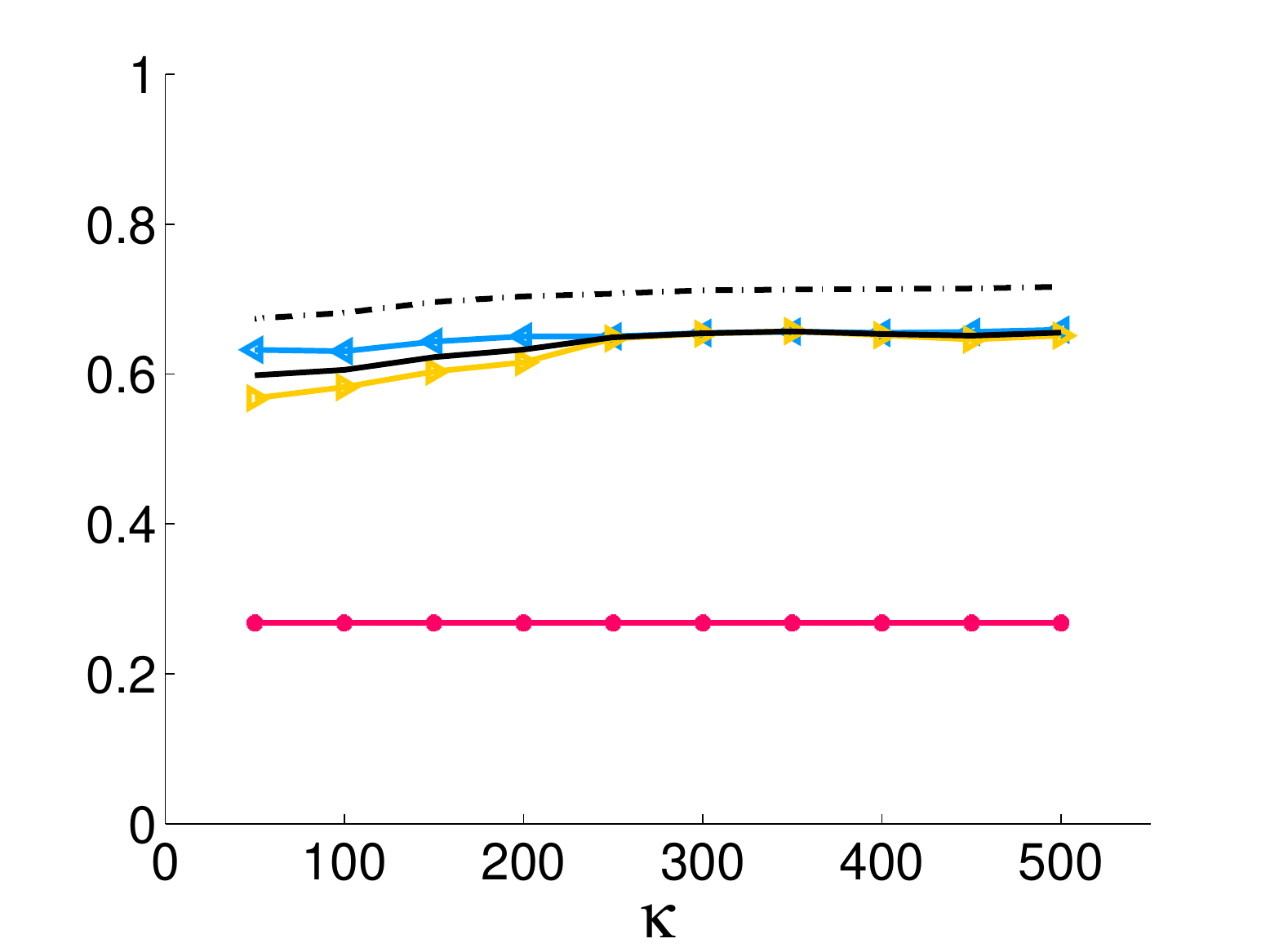}
        %\put(42.5,65){\tiny$\similarity=0.9$}
    \end{overpic}
    \begin{overpic}[width=0.475\linewidth]{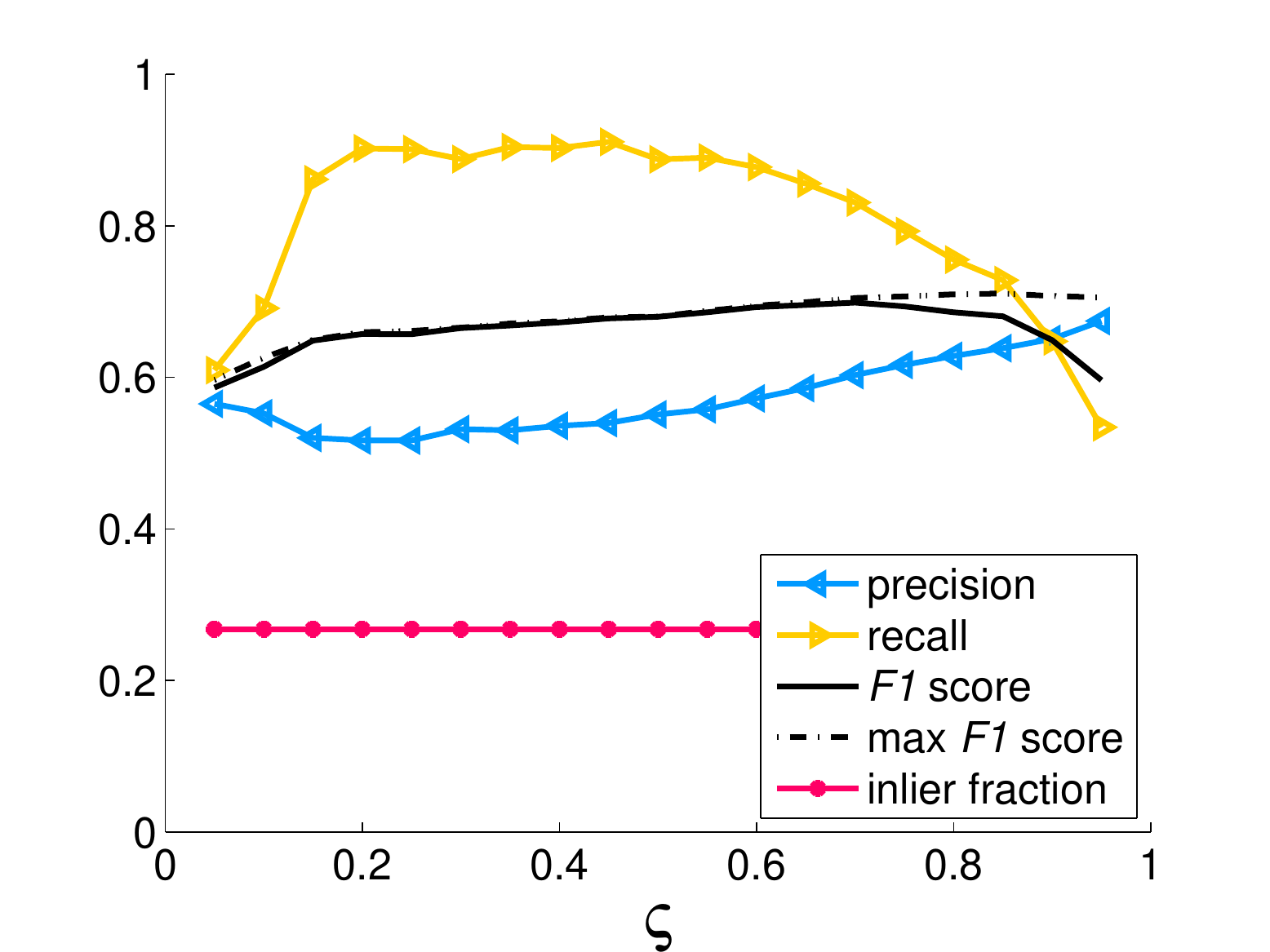}
        %\put(42.5,70){\tiny$\kk=250$}
    \end{overpic}
    \caption{Precision, recall and $F_1$ score at the decision threshold, maximum possible $F_1$ score and inlier fraction for a fixed noise level of $\noise = \SI{2.5}{\milli\meter}$. Left: performance for varying sample counts $\kk$ ($\similarity=0.9$). Right: performance for varying similarity thresholds $\similarity$ ($\kk=250$).}
    \label{fig:output_bunny_sam_sim}
\end{figure}

We also tested the influence of parameter changes at a fixed noise level of $\noise=\SI{2.5}{\milli\meter}$ and performed the matching with a linear change in the sampling size $\kk$ and the similarity $\similarity$ while fixing the other parameter. We show the results in \reffig{fig:output_bunny_sam_sim}, equivalent to the right part of \reffig{fig:output_bunny_noise}.

Interestingly, the sample size $\kk$ has little influence on the results, whereas the similarity $\similarity$ is more crucial to performance. The leftmost plot shows the convergent state of the algorithm for high sample sizes, which we have verified by testing even larger sample sizes. The interpretation of the rightmost plot in \reffig{fig:output_bunny_sam_sim} is that $\similarity$ represents a trade-off between precision and recall. When this value is set high, the method becomes more selective, leading to fewer votes for all correspondences. This gives an increased precision, since the few accepted correspondences are more reliable, but at the expense of recall. Both plots confirm the accuracy of the thresholding method; indeed the $F_1$ scores at the all decision thresholds are very close to the optimal value.

Based on these results, we use $\kk = 250$ and $\similarity = 0.9$ for a good trade-off between speed and accuracy in all the following experiments.

%%%%%%%%%%%%%%%%%%%% BUNNY RESULTS, PARAMETERS

\subsection{Comparative experiments}
In this section, we present comparative experiments carried out on two different datasets. The methods used in the comparison are shortly described below.

\paragraph{$L_2$ distance:}%
The baseline feature distance ranking simply uses the negative of the $L_2$ feature distance \refequ{eq:scoredistance} for ranking correspondences. As noted before, this method is expected to be highly sensitive to \eg repetitive structures.

\paragraph{Ratio:}%
The ratio method \cite{lowe2004} ranks each correspondence by the negative of Lowe's ratio penalty \refequ{eq:scoreratio}. Non-unique feature matches are now removed, but correctness of the remaining matches is not guaranteed.

\paragraph{Geometric consistency:}%
The GC method \cite{chen2007} clusters correspondences by imposing an absolute pairwise distance constraint equal to the Euclidean distance between the feature points. The algorithm initializes a cluster with a seed correspondence and adds all correspondences that are compatible with the seed to the cluster. The clustered correspondences are then marked as visited, and the seed growing repeats until all correspondences have been visited. As an additional step, \cite{aldoma2012} applies RANSAC to each cluster to remove spurious correspondences for increased precision.

For our evaluations, we must apply a proper ranking to the correspondences output by GC. We found that the relative size of the containing cluster to the total input size performs better than \eg the ratio or the RMSE reported by RANSAC. This is intuitive since the cluster size can be seen as an estimate of the inlier fraction, conditioned that a cluster contains inliers.

\subsubsection{Results}
We run comparative tests on two datasets, both synthetic and real. The first dataset of Tombari \etal \cite{tombari2010} consists of 45 synthetic scenes containing between three and five instances of the Stanford models \emph{Armadillo}, \emph{Asian Dragon}, \emph{Bunny}, \emph{Dragon}, \emph{Happy Buddha} and \emph{Thai Statue}. All scenes are contaminated by isotropic Gaussian noise of \SI{10}{\percent} of the spatial resolution, followed by a downsampling to half resolution. The protocol for this dataset is to sample 1000 keypoints per object, which allows us to also test the influence of sparse feature matching on our method. The second dataset by Mian \etal \cite{mian2006} contains four complete 3D models (\emph{Chef}, \emph{Para}, \emph{T-rex} and \emph{Chicken}) and 50 real scenes captured with a laser scanner (see \reffig{fig:corr_results_mian0}). Prior to feature extraction, all models in the laser scanner dataset are downsampled to \SI{2}{\milli\meter}, followed by surface normal estimation \cite{rusu2011}. For all tests, we extract SHOT features and find ground truth inliers using the ground truth poses provided by the datasets, by requiring that matched points must be closer than two resolution units. The mean inlier fraction over all scenes ranges from \SI{1.7}{\percent} (\emph{T-rex}) to \SI{4.0}{\percent} (\emph{Chef}), making the task of finding the inliers very challenging.

\begin{figure}[t]
    \centering
    % STANFORD
    \begin{subfigure}[c]{0.475\linewidth}
        \centering
        \fbox{\includegraphics[height=0.15\linewidth]{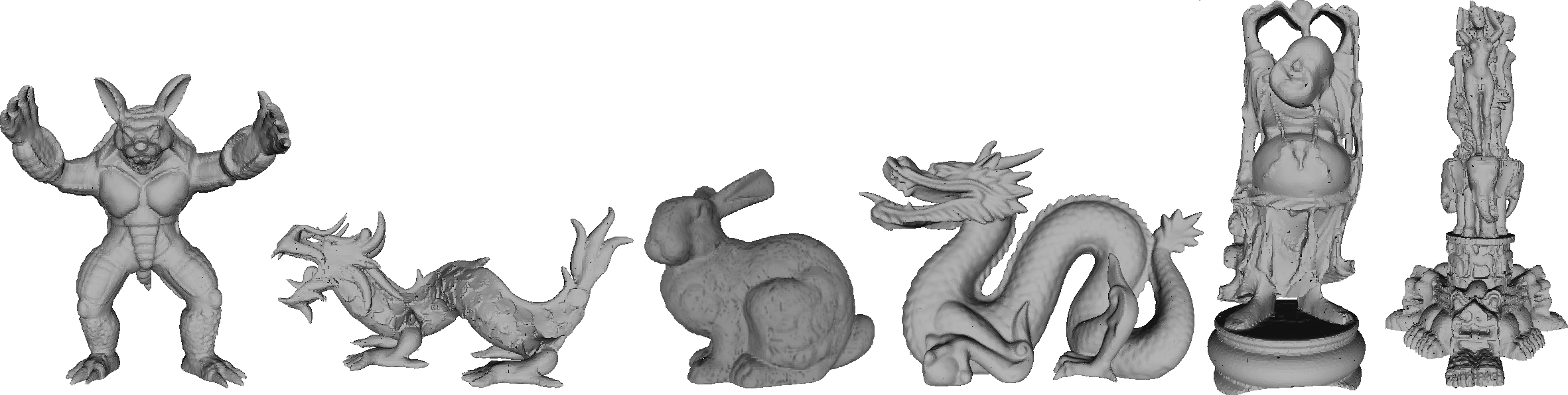}}
        \begin{overpic}[width=\linewidth]{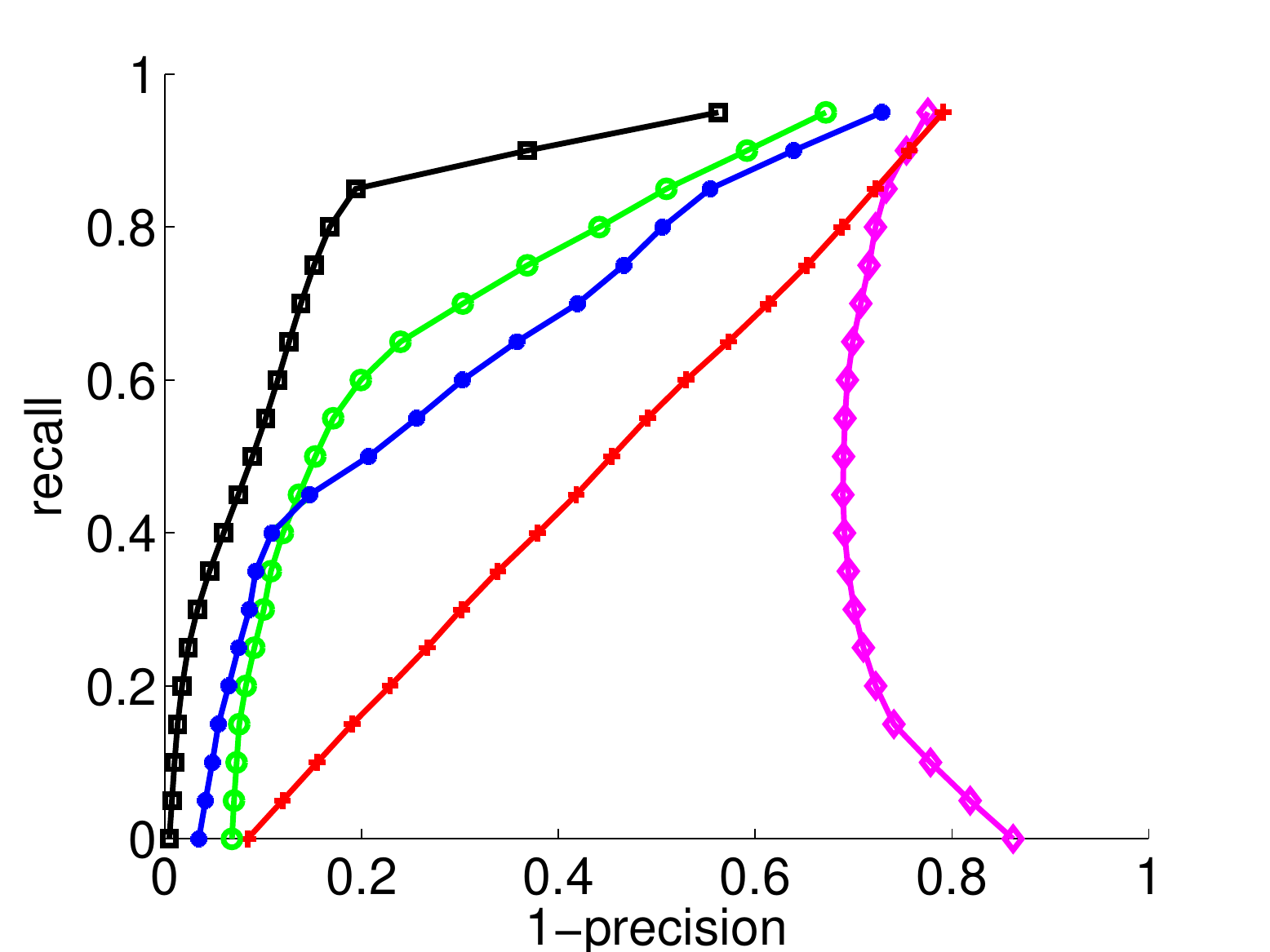}
        \end{overpic}
    \end{subfigure}
    % MIAN
    \begin{subfigure}[c]{0.475\linewidth}
        \centering
        \fbox{\includegraphics[height=0.15\linewidth]{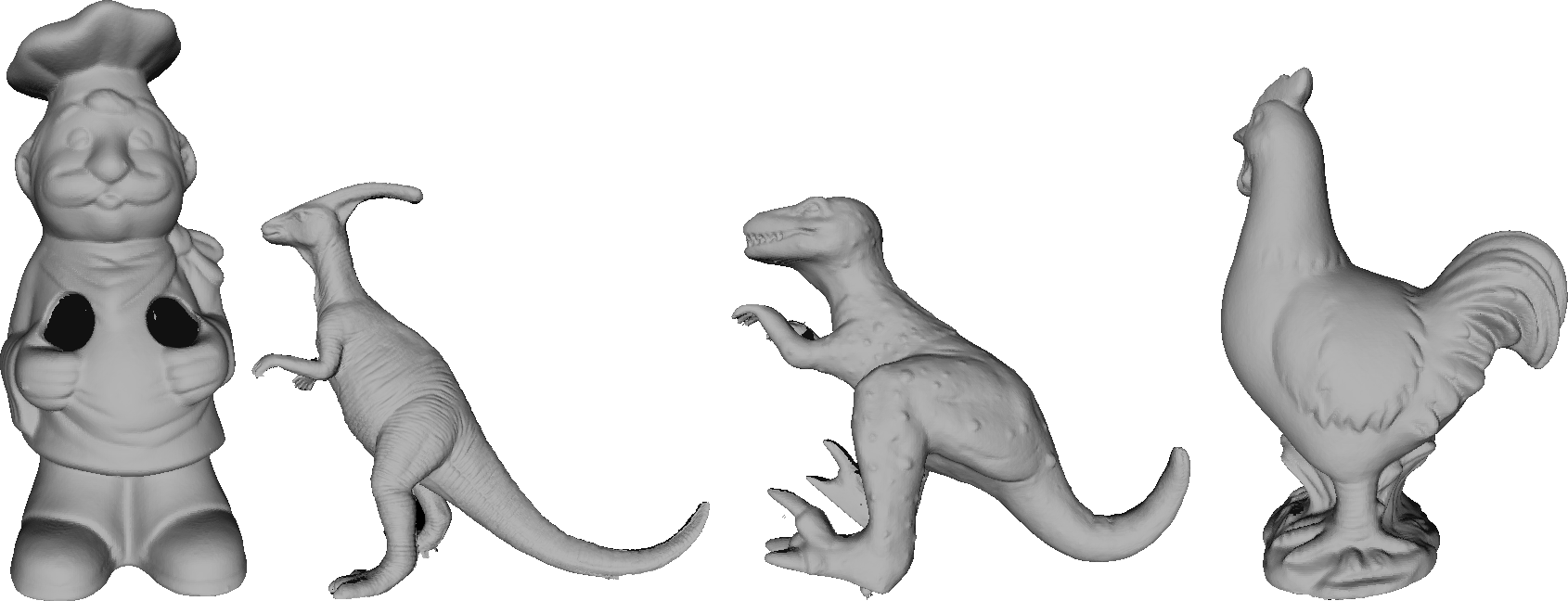}}
        \begin{overpic}[width=\linewidth]{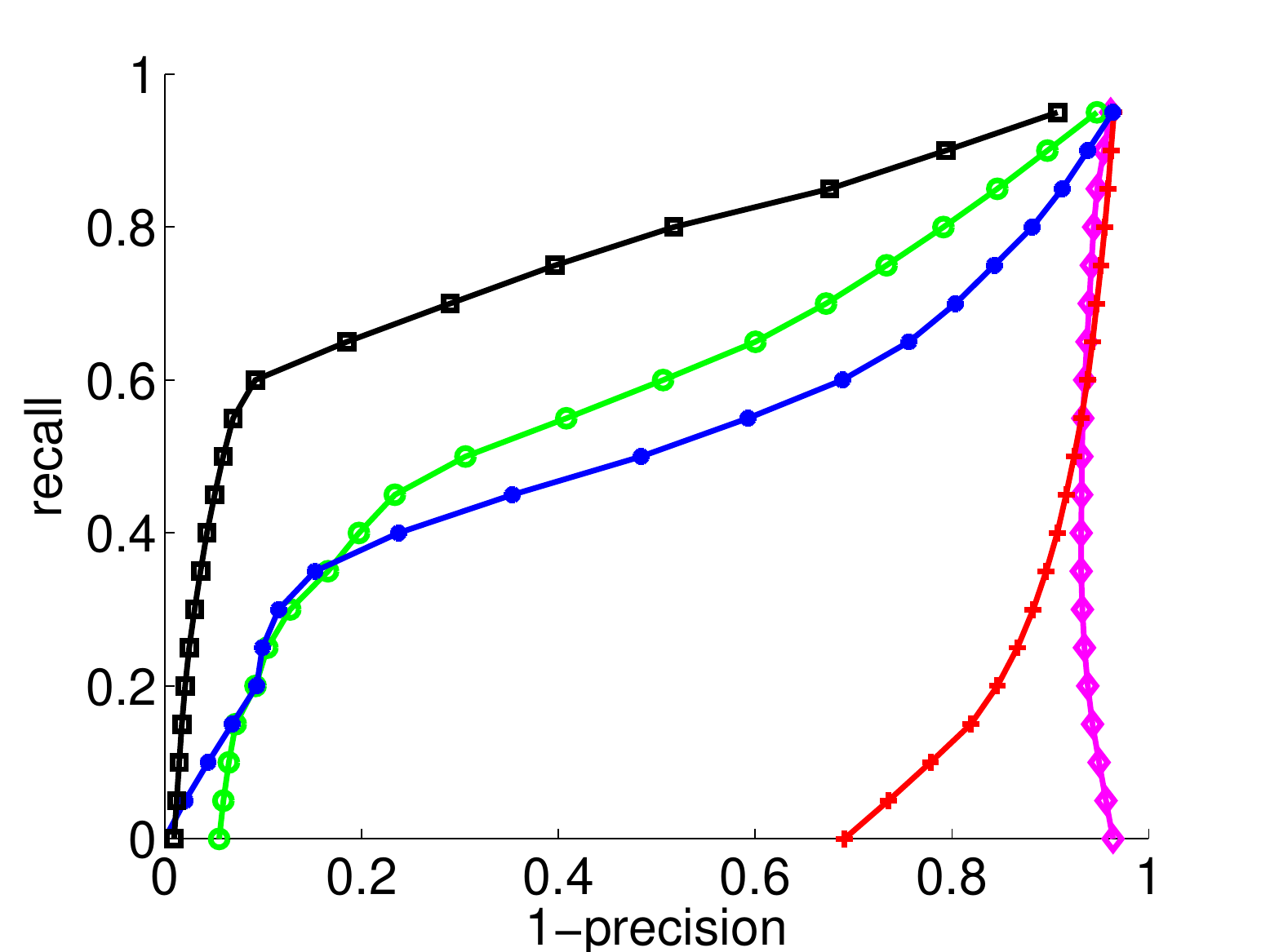}
        \end{overpic}
    \end{subfigure}
    
    \framebox[1.01\width]{\includegraphics[width=0.75\linewidth]{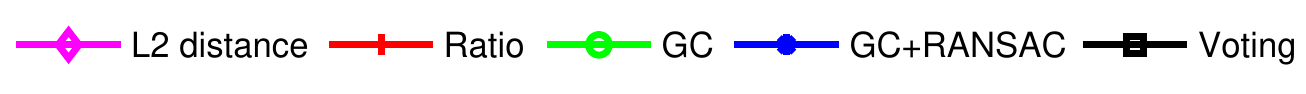}}
    \caption{Overall results for the synthetic feature matching benchmark (left) and the real laser scanner dataset (right).}
    \label{fig:results_shot2_mian}
\end{figure}

Mean \precisionrecall curves for both datasets are reported in \reffig{fig:results_shot2_mian}. As expected, the $L_2$ distance matching shows poor performance. We explain this by the fact that feature distances are very sensitive to repetitive structures. The ratio method shows quite good performance for the synthetic dataset, but quickly degrades to the performance of the distance matching for the real scenes. Surprisingly, GC has better overall performance than GC+RANSAC. However, GC+RANSAC---being more selective---shows a higher initial precision, making it more suitable for algorithms requiring few inliers, \eg pose estimation. The proposed method performs significantly better than all other methods, which we believe comes from the benefit of using different kinds of pairwise geometric constraints.

%%%%%%%%%%%%%%%%%%%% RESULTS, STANFORD AND MIAN

Additional performance measures are reported in \reftab{tab:results_mian}. The feature distance is the baseline required for all methods, and marks the temporal starting point. Since the ratio score only involves a floating point division followed by a subtraction, it is very fast. Compared to the GC methods, our method provides several magnitudes of speedup.

%%%%%%%%%%%%%%%%%%%% ADDITIONAL RESULTS, MIAN
\begin{table}[h]
    \centering
    \begin{footnotesize}
        \begin{tabular}[c]{|l|l|c|c|}
            \hline
            Object         & Method         & Max $F_1$ & Run time [s]  \\
            \hline\hline
            \emph{Chef}    & $L_2$ distance & 0.13      & -             \\
            (28940)        & Ratio          & 0.22      & 0.012         \\
                           & GC             & 0.67      & 1.7           \\
                           & GC+RANSAC      & 0.64      & 1.9           \\
                           & Voting         & 0.85      & 0.33          \\
            \hline
            \emph{Para}    & $L_2$ distance & 0.11      & -             \\
            (16732)        & Ratio          & 0.15      & 0.0058        \\
                           & GC             & 0.57      & 0.64          \\
                           & GC+RANSAC      & 0.54      & 0.77          \\
                           & Voting         & 0.71      & 0.16          \\
            \hline
            \emph{T-rex}   & $L_2$ distance & 0.10      & -             \\
            (15851)        & Ratio          & 0.11      & 0.0051        \\
                           & GC             & 0.56      & 0.56          \\
                           & GC+RANSAC      & 0.47      & 0.65          \\
                           & Voting         & 0.78      & 0.13          \\
            \hline
            \emph{Chicken} & $L_2$ distance & 0.14      & -             \\
            (12324)        & Ratio          & 0.18      & 0.0039        \\
                           & GC             & 0.59      & 0.35          \\
                           & GC+RANSAC      & 0.56      & 0.41          \\
                           & Voting         & 0.80      & 0.11          \\
            \hline
        \end{tabular}
        \caption{Performance measures for the laser scanner dataset. The number below each object is the vertex count. Maximum $F_1$ scores are computed along the mean curves in \reffig{fig:results_shot2_mian}, right. Run times are for full scenes, containing between \approximately 16000 and 23000 vertices.}
        \label{tab:results_mian}
    \end{footnotesize}
\end{table}

\subsection{Application: object detection}
In two final experiments, we demonstrate the power of our method by applying it for object detection. We embed our correspondence matching procedure in a naive detection system as follows. For each object model in the dataset, we input the full set of calculated scene features and calculate feature correspondences $\corrsetf$, now with an increased radius of \SI{0.03}{\meter} for better initial feature matching. Calculating scene features is done once per scene, and takes on average \approximately \SI{1}{\second}. We run our algorithm and take the pose hypothesis of the single top ranked correspondence in the output. We run 10 ICP iterations \cite{besl1992} to refine the result and accept the detection if the aligned object model is covered at least \SI{5}{\percent} by the scene data. We deliberately avoid using sophisticated methods for hypothesis segmentation or cross-verification in order to evaluate the strength of our method alone.

\subsubsection{Results for laser scanner scenes}
We applied the detection method to the real laser scanner dataset, which has been used for object detection comparisons in several works. As shown in \reftab{tab:mian_rec_results}, even with our simplistic system, we achieve good detection performance. Since there are no false positives, precision is \SI{100}{\percent} for all objects. We encourage the reader to compare detection rates, and especially timings, with state of the art recognition systems such as \cite{aldoma2012,drost2010,mian2006,papazov2011}. We believe this demonstrates high potential for the use of our method for higher-level matching tasks such as object detection.

%%%%%%%%%%%%%%%%%%%% DETECTION RESULTS, MIAN
\begin{table}[h]
    \centering
    \begin{footnotesize}
        \begin{tabular}[c]{|l|c|c|}
            \hline
            Object         & Recall [\%] & Time [s] \\
            \hline\hline
            \emph{Chef}    & 100         & 0.39     \\
            \emph{Para}    & 100         & 0.20     \\
            \emph{T-rex}   & 100         & 0.18     \\
            \emph{Chicken} & 90          & 0.15     \\
            \hline
        \end{tabular}
    \end{footnotesize}
    \caption{Detection rates and mean timings for the laser scanner dataset. Timings include both correspondence voting and pose refinement.}
    \label{tab:mian_rec_results}
\end{table}

\subsubsection{Qualitative result from real experiment}
\begin{figure}[t]
    \centering
    \begin{subfigure}[c]{0.35\linewidth}
        \centering
        \fbox{\includegraphics[width=0.5\linewidth]{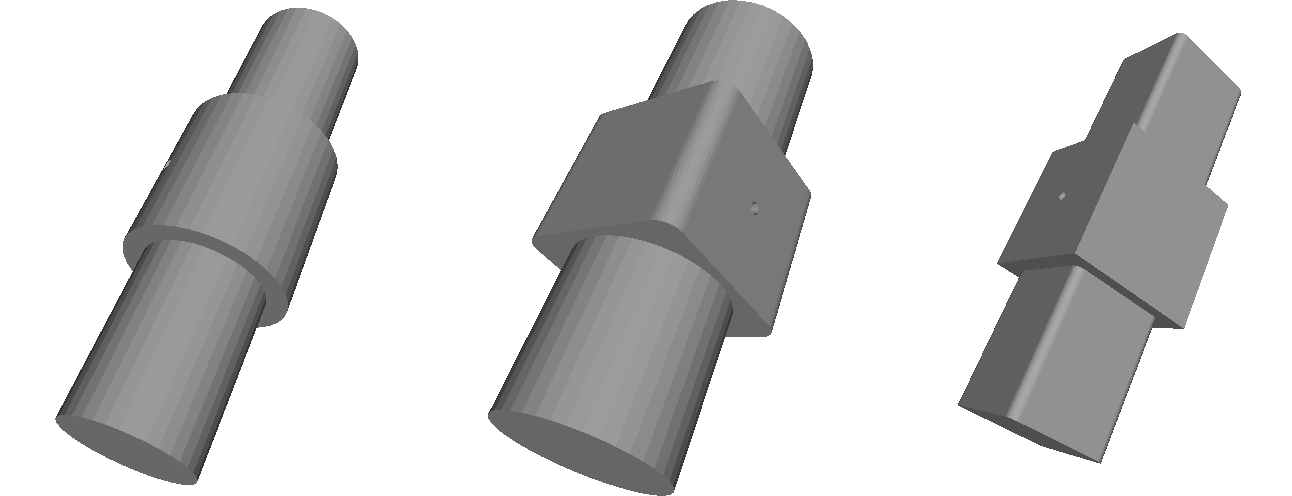}}
        
        \begin{subfigure}[c]{\linewidth}
            \includegraphics[width=\linewidth]{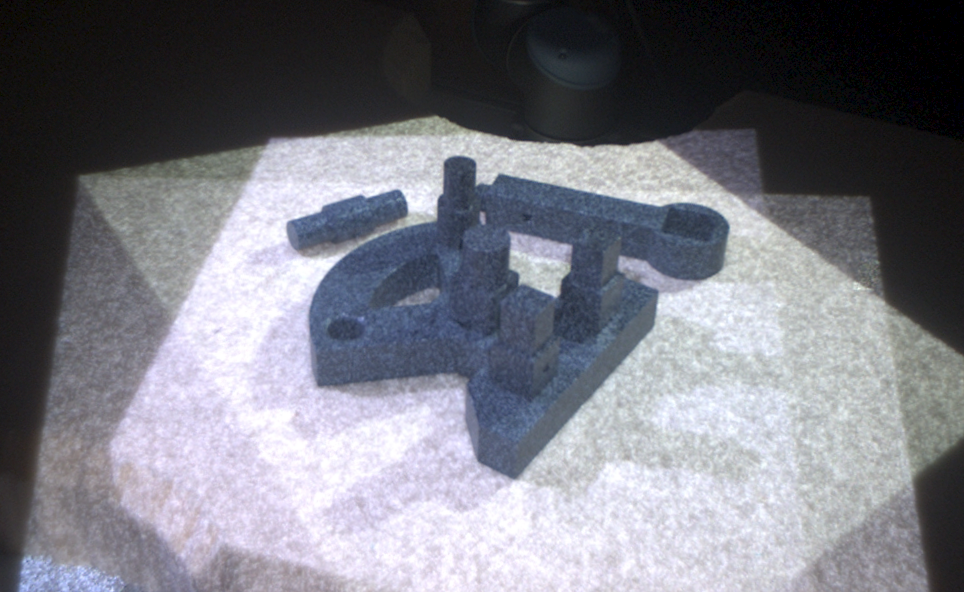}
        \end{subfigure}
    \end{subfigure}
    \begin{subfigure}[c]{0.6\linewidth}
        \centering
        \includegraphics[width=\linewidth]{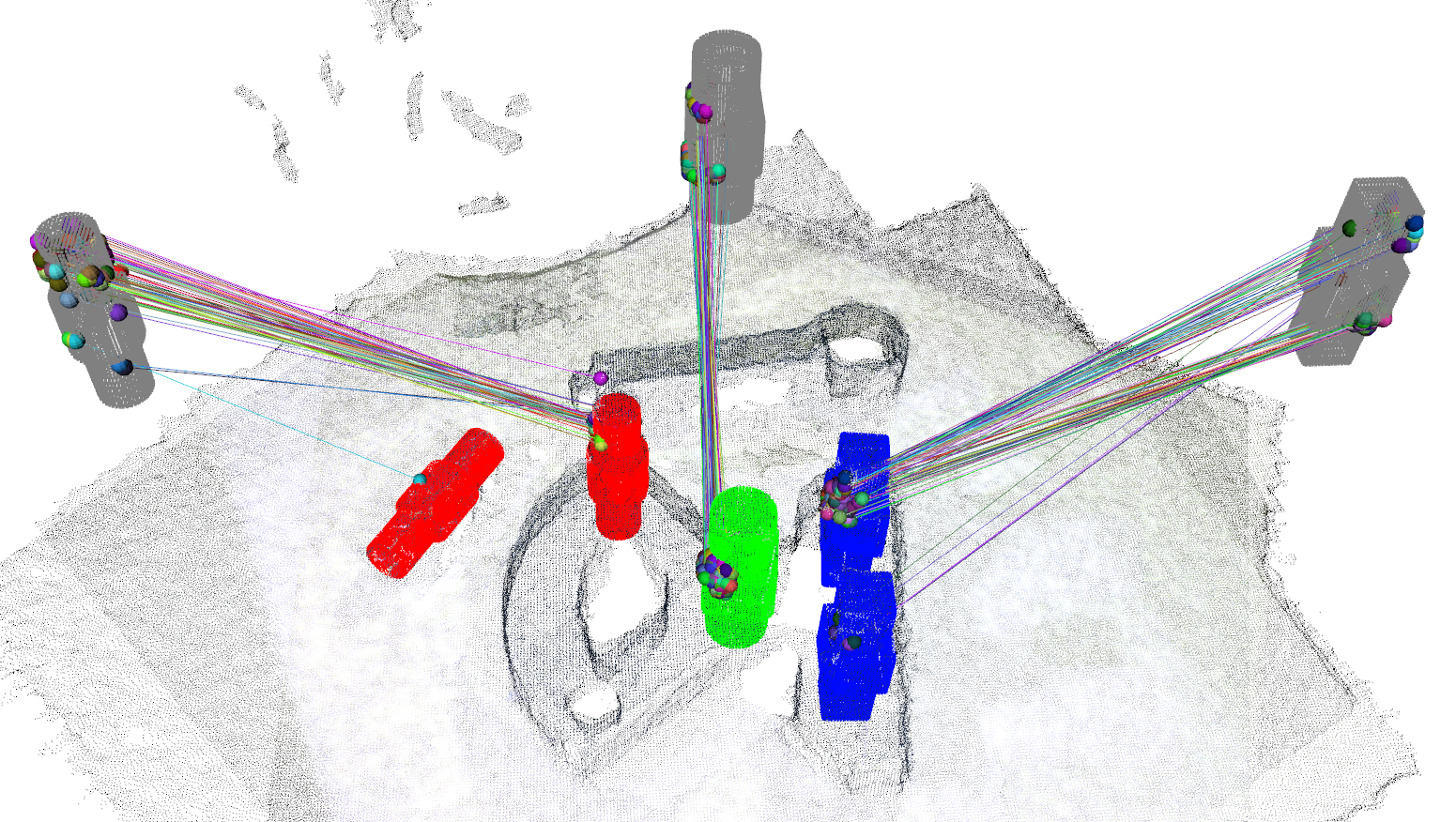}
    \end{subfigure}
    \caption{Left: objects (top) and left frame of one of three texture-projected stereo views used in our setup (bottom). Right: top 100 ranked correspondences and final detections within the point cloud for each object (red: round peg, green: round peg with square handle, blue: square peg).}
    \label{fig:bb_all_detections}
\end{figure}

We finally present a qualitative result from our own experimental setup, consisting of three calibrated stereo cameras. We project texture to the scene before extracting images and performing dense stereo matching. The full point cloud of the scene is obtained by aligning the reconstructed point clouds from the three views, followed by a \SI{2}{\milli\meter} downsampling. The application is robotic (dis)assembly of three very similar pegs, which need to be automatically detected. We perform no scene preprocessing, such as \eg ground plane removal, and input the full point cloud when detecting each object. To allow for multiple instances, we now use the top 100 ranked correspondences per object, and accept the refined pose if it does not overlap with a previous detection of the same object by more than \SI{10}{\percent}. As shown in \reffig{fig:bb_all_detections} on the next page, our method localizes the parts, even when multiple instances are present. The total detection time for this scene, including pose refinements, is \SI{1.9}{\second}.

%%%%%%%%%%%%%%%%%%%% DETECTION RESULTS, MARVIN

%%%%%%%%%%%%%%%%%%%% CONCLUSIONS %%%%%%%%%%%%%%%%%%%%
\section{Conclusions and future work}\label{conclusionsandfuturework}
The method described in this paper allows for efficient and accurate retrieval of correspondences between 3D models based on putative matches obtained by feature matching. Evaluated on different datasets, the proposed method gives an increase in both speed and accuracy by up to several orders of magnitude compared to other methods. We have justified the use of our method for real-life vision problems by testing it for object detection, leading to promising results.

An extension of the method, which we plan to pursue in the future, is multi-instance correspondence voting including several object models. We expect this to achieve sublinear runtime increase in the number of models, which is essential for scalability. By such an extension, we intend to integrate our method into an object recognition framework, allowing for efficient detection of multiple 3D objects.

\paragraph{Acknowledgments:} This work has been supported by the EC project IntellAct (FP7-ICT-269959). The financial support from the The Danish Council for Strategic Research through the project Carmen is gratefully acknowledged.

%%%%%%%%%%%%%%%%%%%% REFERENCES %%%%%%%%%%%%%%%%%%%%
{\small
\bibliographystyle{ieee}
\bibliography{references}
}

\end{document}